\documentclass{article}
\pdfoutput=1
\usepackage{type1cm}        
\usepackage{makeidx}         
\usepackage{graphicx}      
\usepackage{multicol}        
\usepackage[bottom]{footmisc}
\usepackage{newtxtext}       
\usepackage[varvw]{newtxmath}
\usepackage{multirow}
\usepackage{hyperref}\usepackage{breakurl} 
\makeindex  

\begin{document}

\title{Fuzzy Clustering by Hyperbolic Smoothing}
\author{David Mas\'is\thanks{Costa Rica Institute of Technology, Cartago, Costa Rica. E-Mail: \href{mailto: dmasis@itcr.ac.cr}{dmasis@itcr.ac.cr}} 
\and 
Esteban Segura\thanks{CIMPA \& School of Mathematics, University of Costa Rica, San Jos\'e, Costa Rica. E-Mail: estebaseguraugalde@ucr.ac.cr}
\and
Javier Trejos\thanks{CIMPA \& School of Mathematics, University of Costa Rica, San Jos\'e, Costa Rica. E-Mail: javier.trejos@ucr.ac.cr} 
\and 
Adilson Xavier\thanks{ Universidade Federal de Rio de Janeiro, Brazil. E-Mail: adilson.xavier@gmail.com}
}

\maketitle

\abstract{We propose a novel method for building fuzzy clusters of large data sets, using a smoothing numerical approach.
The usual sum-of-squares criterion is relaxed so the search for good fuzzy partitions is made on a continuous space, rather than a combinatorial space as in classical methods \cite{Hartigan}. The smoothing allows a conversion from a strongly non-differentiable problem into differentiable subproblems of optimization without constraints of low dimension, by using a differentiable function of infinite class.
For the implementation of the algorithm we used the statistical software $R$ and the results obtained were compared to the traditional fuzzy $C$--means method, proposed by Bezdek \cite{Bezdek}.}
\medskip


\section{Introduction}

Methods for making groups from data sets are usually based on the idea of disjoint sets, such as the classical crisp clustering. The most well known are hierarchical and $k$-means \cite{Hartigan}, whose resulting clusters are sets will no intersection. However, this restriction may not be natural for some applications, where the condition for some objects may be to belong to two or more clusters, rather than only one.
Several methods for constructing overlapping clusters have been proposed in the literature \cite{Pyramids,Dunn, Hartigan}.
Since Zadeh introduced the concept of fuzzy sets \cite{Zadeh}, the principle of belonging to several clusters has been used in the sense of a degree of membership to such clusters.
In this direction, Bezdek \cite{Bezdek} introduced a fuzzy clustering method that became very popular since it solved the problem of representation of clusters with centroids and the assignment of objects to clusters, by the minimization of a well-stated numerical criterion.
Several methods for fuzzy clustering have been proposed in the literature; a survey of these methods can be found in \cite{Yang}.

In this paper we propose a new fuzzy clustering method based on the numerical principle of hyperbolic smoothing \cite{Xav}. 
Fuzzy $C$-Means method is presented in Section \ref{sec:FCM}
and our proposed Hyperbolic Smoothing Fuzzy Clustering method in Section \ref{sec:HSFC}.
Comparative results between these two methods are presented  in Section \ref{sec:results}.
Finally, Section \ref{sec:conclusion} is devoted to the concluding remarks.

\section{Fuzzy Clustering}
\label{sec:FCM}

The most well known method for fuzzy clustering is the original Bezdek's $C$-means method \cite{Bezdek} and
it is based on the same principles of $k$-means or dynamical clusters  \cite{Bock}, that is, iterations on two main steps: i) class representations by the optimization of a numerical criterion, and ii) assignment to the closest class representative in order to construct clusters; these iterations are made until a convergence is reached to a local minimum of the overall quality criterion.

Let us introduce the notation that will be used and the numerical criterion for optimization.
Let $\textbf{X}$ be a  $n \times p$ data matrix containing $p$ numerical observations over $n$ objects.
We look for a $K\times p$ matrix $\textbf{G}$ that represents centroids of $K$ clusters of the $n$ objects and an $n\times K$ membership matrix with elements $\mu_{ik}\in[0,1]$, such that the following criterion is minimized:
\begin{eqnarray} 
\begin{array}{ll}
\multicolumn{2}{c}{W(\textbf{X}, \textbf{U},\textbf{G})=\displaystyle \sum_{i=1}^{n}\,\displaystyle \sum_{k=1}^{K}\, (\mu_{ik})^{m}\; \Vert\textbf{x}_{i}-\textbf{g}_{k}\Vert^{2}}\\
\mbox{ subject to } & \sum_{k=1}^{K}\,\mu_{ik}=1, \mbox{ for all } i\in \{1,2,\ldots,n\}\\
& 0< \sum_{i=1}^{n}\,\mu_{ik}<n, \mbox{ for all }k\in \{1,2,\ldots,K\},
\end{array}
\label{funcional}
\end{eqnarray}
where $\textbf{x}_{i}$ is the $i$-th row of $\textbf{X}$  and $\textbf{g}_{k}$ is the $k$-th row of $\textbf{G}$, representing in $\mathbb{R}^{p}$ the centroid of the $k$-th cluster. 

The parameter $m\neq 1$  in (\ref{funcional}) controls the fuzzyness of the clusters. According to the literature \cite{Yang}, it is usual to take $m=2$, since greater values of $m$ tend to give very low values of $\mu_{ik}$, tending to the usual crisp partitions such as in $k$-means. We also assume that the number of clusters, $K$, is fixed.

Minimization of (\ref{funcional}) represents a non linear optimization problem with constraints, which can be solved using Lagrange multipliers as presented in \cite{Bezdek}. The solution, for each row of the centroids matrix, given a matrix $\textbf{U}$, is:
\begin{equation}
\textbf{g}_{k}= { \sum_{i=1}^{n}\,(\mu_{ik})^{m}\textbf{x}_{i}} \left/ {\displaystyle \sum_{i=1}^{n}\,(\mu_{ik})^{m}}\right..
\label{centroids}
\end{equation}
The solution for the 
membership matrix, given a matrix centroids $\textbf{G}$, is \cite{Bezdek}:
\begin{equation}
\mu_{ik}=\left[ \sum_{j=1}^{K}\,\left( \frac{||\textbf{x}_{i}-\textbf{g}_{k}||^{2}}{||\textbf{x}_{i}-\textbf{g}_{j}||^{2}}\right)^{1/(m-1)}\right]^{-1}.
\label{membership}
\end{equation}

The following pseudo-code shows the mains steps of Bezdek's Fuzzy $C$-Means method \cite{Bezdek}.

\paragraph{Bezdek's Fuzzy c-Means (FCM) Algorithm}
\begin{enumerate}
\item Initialize fuzzy membership matrix $\textbf{U}=[\mu_{ik}]_{n\times K}$\; 	
\item Compute centroids for fuzzy clusters according to (\ref{centroids})
\item Update membership matrix $\textbf{U}$ according to (\ref{membership})
\item  If improvement in the criterion is less than a threshold, then stop; otherwise go to Step 2.
\end{enumerate}

Fuzzy $C$-Means method starts from an initial partition that is improved in each iteration, according to (\ref{funcional}), applying Steps 2 and 3 of the algorithm. It is clear that this procedure may lead to local optima of (\ref{funcional}) since iterative improvement in (\ref{centroids}) and (\ref{membership}) is made by a local search strategy.

\section{Algorithm for Hyperbolic Smoothing Fuzzy Clustering}
\label{sec:HSFC}

For the clustering problem of the $n$ rows of data matrix \textbf{X} in $K$ clusters, we can seek for the minimum distance between every $\textbf{x}_{i}$ and its class center $\textbf{g}_{k}$:
\begin{equation*}
z_{i}^2=\displaystyle \min_{\textbf{g}_{k}\in \textbf{G}}\Vert\textbf{x}_{i}-\textbf{g}_{k}\Vert^2_{2} \label{zi}
\end{equation*}
where $\Vert\cdot\Vert_{2}$ is the Euclidean norm.
The minimization can be stated as a sum-of-squares:
\begin{equation*}
\displaystyle \min  \sum_{i=1}^{n}\,  \min_{\textbf{g}_{k}\in \textbf{G}} \Vert\textbf{x}_{i}-\textbf{g}_{k}\Vert_{2}^{2}= \min  \sum_{i=1}^{n}\,z_{i}^{2}
\end{equation*}
leading to the following constrained problem:
$$
\min\displaystyle \sum_{i=1}^{n}\,z_{i}^{2} \mbox{ subject to } z_{i}=\displaystyle \min_{\textbf{g}_{k}\in \textbf{G}}\Vert\textbf{x}_{i}-\textbf{g}_{k}\Vert_{2}, \mbox{ with } i=1,\ldots,n.
$$
\pagebreak

\noindent
This is equivalent to the following minimization problem:
$$
\min\displaystyle \sum_{i=1}^{n}\,z_{i}^{2} \mbox{ subject to } z_{i}-\Vert\textbf{x}_{i}-\textbf{g}_{k}\Vert_{2}\leq 0 ,\mbox{ with } i=1,\ldots,n \mbox{ and } k=1,\ldots,K.
$$
Considering the function: $\varphi(y)=\max(0,y)$, we obtain the problem:
$$
\min\displaystyle \sum_{i=1}^{n}\,z_{i}^{2} \mbox{ subject to } \displaystyle \sum_{k=1}^{K}\, \varphi(z_{i}-\Vert\textbf{x}_{i}-\textbf{g}_{k}\Vert_{2})=0\mbox{ for }i=1,\ldots,n. 
$$
That problem can be re-stated as the following one:
$$
\min\displaystyle \sum_{i=1}^{n}\,z_{i}^{2} \mbox{ subject to } \displaystyle \sum_{k=1}^{K}\, \varphi(z_{i}-\Vert\textbf{x}_{i}-\textbf{g}_{k}\Vert_{2})>0,\mbox{ for }i=1,\ldots,n.
$$
Given a perturbation $\epsilon>0$ it leads to the problem:
$$
\min\displaystyle \sum_{i=1}^{n}\,z_{i}^{2} \mbox{ subject to }\displaystyle \sum_{k=1}^{K}\, \varphi(z_{i}-\Vert\textbf{x}_{i}-\textbf{g}_{k}\Vert_{2})\geq \epsilon\mbox{ for }i=1,\ldots,n.
$$
It should be noted that function $\varphi$ is not differentiable. Therefore, we will make a smoothing procedure in order to formulate a differentiable function and proceed with a minimization by a numerical method. For that, consider the function:
$
\psi(y,\tau)= \frac{y+\sqrt{y^{2}+\tau^{2}}}{2},
$ 
for all $y\in \mathbb{R}$, $\tau>0$, and the function:
	$\theta(\textbf{x}_{i},\textbf{g}_{k},\gamma)=\sqrt{ \sum_{j=1}^{p}\,(x_{ij}-g_{kj})^{2}+\gamma^{2}}$,
for $\gamma>0$.
Hence, the minimization problem is transformed into:
$$
\min\displaystyle \sum_{i=1}^{n}\,z_{i}^{2} \mbox{ subject to } \displaystyle \sum_{k=1}^{K}\, \psi(z_{i}-\theta(\textbf{x}_{i},\textbf{g}_{k},\gamma),\tau)\geq \epsilon, \mbox{ for } i=1,\ldots,n.
$$

Finally, according to the Karush--Kuhn--Tucker conditions \cite{Kar, KT}, all the constraints are active and the final formulation of the problem is:
\begin{equation}
\begin{array}{ll}
& \min\displaystyle \sum_{i=1}^{n}\,z_{i}^{2}\\
\mbox{ subject to } & h_{i}(z_{i},\textbf{G})= \displaystyle\sum_{k=1}^{K}\, \psi(z_{i}-\Vert\textbf{x}_{i}-\textbf{g}_{k}\Vert_{2},\tau)-\epsilon=0, \mbox{ for } i=1,\ldots,n,\\
& \epsilon,\tau,\gamma>0.
\end{array}
\label{problem}
\end{equation}
Considering (\ref{problem}), in \cite{Xav} it was stated the Hyperbolic Smoothing Clustering Method presented in the following algorithm.

\paragraph{Hyperbolic Smoothing Clustering Method (HSCM) Algorithm}
\begin{enumerate}
\item Initialize cluster membership matrix $\textbf{U}=[\mu_{ik}]_{n\times K}$
\item Choose initial values: $\textbf{G}^{0}, \gamma^{1}, \tau^{1}, \epsilon^{1}$
\item Choose values: $0<\rho_{1}<1$, $0<\rho_{2}<1$, $0<\rho_{3}<1$
\item Let $l=1$
\item Repeat steps 6 and 7 until a stop condition is reached:
\item Solve problem (P): $\min f(\textbf{G})=\displaystyle \sum_{i=1}^{n}\,z_{i}^{2}$  with $\gamma=\gamma^{l}$, $\tau=\tau^{l}$ y $\epsilon=\epsilon^{l}$, $\textbf{G}^{l-1}$ being the initial value and $\textbf{G}^{l}$ the obtained solution
\item Let $\gamma^{l+1}=\rho_{1}\gamma^{l}$,\; $\tau^{l+1}=\rho_{2}\tau^{l}$,\; $\epsilon^{l+1}=\rho_{3}\epsilon^{l}$ y $l=l+1$.
\end{enumerate}

The most relevant task in the hyperbolic smoothing clustering method is finding the zeroes of the function 
	$h_{i}(z_{i},\textbf{G})= \sum_{k=1}^{K}\, \psi(z_{i}-\Vert\textbf{x}_{i}-\textbf{g}_{k}\Vert_{2},\tau)-\epsilon=0$
%
for $i=1,\ldots,n$.
In this paper, we used the Newton-Raphson method for finding these zeroes \cite{Burden}, particularly the BFGS procedure \cite{Li}.
Convergence of the Newton-Raphson method was successful, mainly, thank to a good choice of initial solutions. 
In our implementation, these initial approximations were generated by calculating the minimum distance between the $i$-th object and the $k$-th centroid for a given partition.
Once the zeroes $z_{i}$ of the functions  $h_{i}$ are obtained, it is implemented the hyperbolic smoothing.
The final solution for this method consists on solving a finite number of optimization subproblems corresponding to problem (P) in Step 6 of the HSCM algorithm.
Each one of these subproblems was solved with the R routine  \textit{optim} \cite{R}, a useful tool for solving optimization problems in non linear programming.
As far as we know there is no closed solution for solving this step. For the future, we can consider writing a program by our means, but for this paper we are using this R routine.

Since we have that: $\sum_{k=1}^{K}\, \psi(z_{i}-\theta(\textbf{x}_{i},\textbf{g}_{k},\gamma),\tau)=\epsilon$, then each entry  $\mu_{ik}$ of the membership matrix is given by:
$\mu_{ik}=\frac{\psi(z_{i}-d_{k},\tau)}{\epsilon}.$
It is worth to note that fuzzyness is controlled by parameter $\epsilon$.

The following algorithm contains the main steps of the Hyperbolic Smoothing Fuzzy Clustering (HSFC) method.

\paragraph{Hyperbolic Smoothing Fuzzy Clustering (HSFC) Algorithm}
\begin{enumerate}
		\item Set $\epsilon>0$
		\item Choose initial values for: $\textbf{G}^{0}$ (centroids matrix), $\gamma^{1}$, $\tau^{1}$ y $N$ (maximum number of iterations)
		\item Choose values: $0<\rho_{1}<1$,\; $0<\rho_{2}<1$
		\item Set $l=1$
		\item While $l\leq N$:
		\item Solve the problem  (P): Minimize $f(\textbf{G})= \sum_{i=1}^{n}\,z_{i}^{2}$  con $\gamma=\gamma^{(l)}$ y $\tau=\tau^{(l)}$, with an initial point $\textbf{G}^{(l-1)}$ and $\textbf{G}^{(l)}$ being the obtained solution
		\item Set $\gamma^{(l+1)}=\rho_{1}\gamma^{(l)}$,$\tau^{(l+1)}=\rho_{2}\tau^{(l)}$, y $l=l+1$
		\item Set $\mu_{ik}=\psi(z_{i}-\theta(\textbf{x}_{i},\textbf{g}_{k},\gamma),\tau)/\epsilon$ para $i=1,\ldots,n$ y $k=1,\ldots,K$.
\end{enumerate}

\section{Comparative Results}
\label{sec:results}

Performance of the HSFC method was studied on a data table well known from the literature, the Fisher's iris \cite{Fisher} and 16 simulated data tables built from a semi-Monte Carlo procedure \cite{PSOclus}. 

For comparing FCM and HSFC, we used the implementation of FCM in R package \textit{fclust} \cite{Gio}.
This comparison was made upon the within class sum-of-squares:\linebreak
$W(P)= \sum_{k=1}^{K}\, \sum_{i=1}^{n}\,\mu_{ik}\|\textbf{x}_{i}-\textbf{g}_{k}\|^{2}$.
Both methods were applied 50 times and the best value of $W$ is reported.
For simplicity here, for HSFC we used the following parameters:
$\rho_{1}=\rho_{2}=\rho_{3}=0.25$, $\epsilon=0.01$ and $\gamma=\tau=0.001$ as initial values. 
In Table \ref{tab:classic:tables} the results for Fisher's iris are shown, in which case HSFC performs slightly better. It contains the  Adjusted Rand Index (ARI) \cite{ARI} between HSFC and the best FCM result among 100 runs; RI and ARI compare fuzzy membership matrices crisped into hard partitions.

\begin{table}
	\centering
\caption{Minimum sum-of-squares (SS) reported for the Fisher's iris data table with HSFC and FCM, $K$ being the number of clusters, RI and ARI comparing both methods. In bold best method.}
\label{tab:classic:tables}
	\begin{tabular}{p{20mm}|p{5mm}p{20mm}p{20mm}p{8mm}} 
\hline\noalign{\smallskip}
Table & $K$ & SS for HSFC & SS for FCM &  ARI\\ 
\noalign{\smallskip}\hline\noalign{\smallskip}
		& 2 &  \textbf{152.348}  & 152.3615  & \multicolumn{1}{c}{1}\\
{Fisher's iris}       & 3 &  \textbf{78.85567}  & 78.86733 & 0.994\\
		& 4 &  57.26934 & 57.26934 & 0.980\\
\noalign{\smallskip}\hline\noalign{\smallskip}
	\end{tabular}
\end{table}

Simulated data tables were generated in a controlled experiment as in \cite{PSOclus}, with random numbers following a Gaussian distribution. Factors of the experiment were:
\vspace{-0.2cm}
\begin{itemize}
\item The number of objects (with 2 levels, $n=105$ and $n=525$).
\item The number of clusters (with levels $K=3$ and $K=7$).
\item Cardinality (card) of clusters, with levels i) all with the same number of objects (coded as card($=$)), and ii) one large cluster with 50\% of objects and the rest with the same number (coded as card($\not=$)).
\item Standard deviation of clusters, with levels i) all  Gaussian random variables with standard deviation (SD) equal to one (coded as SD($=$)), and ii) one cluster with  SD=3 and the rest with SD=1 (coded as SD($\not=$)).
\end{itemize}
Table \ref{table:nombre} contains codes for simulated data tables according to the codes we used.

\begin{table}
\centering
\caption{Codes and characteristics of simulated data tables; $n$: number of objects, $K$: number of clusters, card: cardinality, DS: standard deviation.}\label{table:nombre}
\begin{tabular}{lp{5cm}|lp{4.1cm}} 
\hline\noalign{\smallskip}
{Table}   & {Characteristcs} & {Table} & {Characteristcs} \\ 
\noalign{\smallskip}\hline\noalign{\smallskip}
T1 & $n=525$, $K=3$, card($=$), SD($=$)     & 
T9   & $n=525$, $K=3$, card($\not=$), DS($=$) \\ 
T2 & $n=525$, $K=7$, card($=$), SD($=$)     & 
T10  & $n=525$, $K=7$, card($\not=$), DS($=$) \\ 
T3 & $n=105$, $K=3$, card($=$), SD($=$)     & 
T11  & $n=105$, $K=3$, card($\not=$), DS($=$) \\ 
T4 & $n=105$, $K=7$, card($=$), SD($=$)     & 
T12  & $n=105$, $K=7$, card($\not=$), DS($=$) \\
T5 & $n=525$, $K=3$, card($=$), SD($\not=$) & 
T13  & $n=525$, $K=3$, card($\not=$), DS($\not=$) \\
T6 & $n=525$, $K=7$, card($=$), SD($\not=$) & 
T14  & $n=525$, $K=7$, card($\not=$), DS($\not=$)\\
T7 & $n=105$, $K=3$, card($=$), SD($\not=$) & 
T15  & $n=105$, $K=3$, card($\not=$), DS($\not=$)\\
T8 & $n=105$, $K=7$, card($=$), SD($\not=$) & 
T16  & $n=105$, $K=7$, card($\not=$), DS($\not=$) \\
\noalign{\smallskip}\hline\noalign{\smallskip}
	\end{tabular}
\end{table}

Table \ref{table:results} contains the minimum values of the sum-of-squares obtained for our HSFC and Bezdek's FCM methods; the best solution of 100 random applications for FCM in presented and one run of HSFC. It also contains the ARI values for comparing HSFC solution with that best solution of FCM.
It can be seen that, generally, HSFC method tends to obtain better results than FCM, with only few exceptions. In 23 cases HSFC obtains better results, FCM is better in 5 cases, and results are in same in 17 cases. However, ARI shows that partitions tend to be very similar with both methods.

\begin{table}
	\centering
\caption{Minimum sum-of-squares (SS) reported for HSFC and FCM methods on the simulated data tables.
Best method in bold.}
\label{table:results}
{\small
\begin{tabular}{p{7mm}p{3mm}p{12mm}p{12mm}p{18mm}|p{7mm}p{3mm}p{12mm}p{12mm}p{12mm}} 
\hline\noalign{\smallskip}
Table   & $K$ &  SS for & SS for  & ARI &
Table   & $K$ &  SS for & SS for & ARI\\
   &  &  HSFC & FCM  & &
   &  &  HSFC & FCM  &\\
\noalign{\smallskip}\hline\noalign{\smallskip}
		& 2 & \textbf{7073.402} & 7073.814 & 0.780 & & 2 & 12524.31 & 12524.31 & 0.900\\ 
T1      & 3 & 3146.119 & 3146.119  & 1 &
T9  & 3 & \textbf{9269.361} & 9269.611 & 1\\
		& 4 & 2983.651 & 2983.651  & 1 &	 & 4 & 6298.47 & \textbf{6298.368} & 1 \\ \hline
		
		& 2 & \textbf{16987.19} & 16987.71  &  0.764 &   & 2 & \textbf{5466.893} & 5466.912 & 0.890\\
T2      & 3 & 11653.22 & 11653.22  & 1 &
T10  & 3 & 2977.58 & 2977.58 & 1\\
		& 4 & \textbf{7776.855} & 7777.396  & 1 & & 4 & \textbf{2745.721} & 2746.671 & 1  \\ \hline
		
		& 2 & \textbf{3923.051} & 3923.062  &     0.763 & & 2 & \textbf{2969.247} & 2969.32  &  0.860\\ 
T3      & 3 & 2917.13  & 2917.13   & 0.754 &
T11  & 3 & 1912.323 & 1912.323  & 1\\
		& 4 & 2287.523 & \textbf{2256.298}  &  0.993 &   & 4 & 1401.394 & 1401.394 & 1\\ \hline
		
		& 2 & \textbf{1720.365} & 1720.374  &  0.992 &   & 2 & 1816.056 & 1816.056 & 1\\
T4      & 3 & 569.3112 & 569.3112  & 1 &
T12  & 3 & 525.7118 & 525.7118 & 1 \\
		& 4 & 535.5491 & \textbf{535.3541}  &  1 &   & 4 & \textbf{477.0593} & 477.2696 & 1\\ \hline
		
		& 2 & 15595.67 & 15595.67  &  0.910 	& & 2 & \textbf{12804.03} & 12805.05 & 0.920 \\ 
T5      & 3 & \textbf{11724.93} & 11725.28  & 1 &
T13  & 3 & \textbf{8816.805} & 8817.702 & 1\\
		& 4 & 8409.738 & 8409.738  & 0.984 &    & 4 & \textbf{6293.774} & 6293.951 & 1\\ \hline
		
		& 2 & 11877.96 & 11877.96  & 0.970   &  & 2 & \textbf{16228.07} & 16228.98  & 0.920\\
T6      & 3 & \textbf{8299.779} & 8300.718  & 1 &
T14  & 3 & \textbf{7255.113} & 7255.423 & 1\\
		& 4 & \textbf{7212.611} & 7213.725  & 1    &  & 4 & 6427.313 & 6427.313 & 1\\ \hline
		
		& 2 & \textbf{4336.261} & 4336.507  &  0.955 &	 & 2 & \textbf{2616.286} & 2616.943  & 1 \\ 
T7      & 3 & 3041.076 & 3041.076  & 1 &
T15  & 3 & \textbf{1978.017} & 1978.233 & 1\\
		& 4 & \textbf{2395.683} & 2421.333  & 1 	&  & 4 & \textbf{1526.895} & 1526.953 & 1 \\ \hline
		
		& 2 & 1767.43  & 1767.43 & 1 &   & 2 & 2226.923 & \textbf{2226.212}  &  0.962  \\
T8      & 3 & \textbf{1380.766} & 1381.019  & 1 &
T16  & 3 & \textbf{1232.074} & 1232.124 & 1  \\
		& 4 & 1215.302 & \textbf{1211.235}  &  1 &   & 4 & \textbf{982.7074} & 982.9721 & 1\\ 
\noalign{\smallskip}\hline\noalign{\smallskip}
	\end{tabular}
}
\end{table}

\section{Concluding Remarks}
\label{sec:conclusion}

In hyperbolic  smoothing, parameters $\tau$, $\gamma$ and $\epsilon$ tend to zero, so the constraints in the subproblems make that problem (P) tends to solve (\ref{funcional}).
Parameter $\epsilon$ controls the fuzzyness degree in clustering; the higher it is, the solution becomes more and more fuzzy; the less it is, the clustering is more and more crisp.
In order to compare results and efficiency of the HSFC method, zeroes of functions $h_{i}$ can be obtained with any method  for solving equations in one variable or a predefined routine.
According to the results we obtained so far and the implementation of the hyperbolic smoothing for fuzzy clustering, we can conclude that, generally, the HSFC method has a slightly better performance than original Bezdek's FCM on small real and simulated data tables.
Further research is required for testing performance of HSFC method on very large data sets, with measures of efficiency, quality of solutions and running time.
We are also considering to study further comparisons between HSFC and FCM with different indices, and writing the program for solving Step 6 in HSFC algorithm, that is the minimization of $f(G)$, by our means, instead of using the \textit{optim} routine in R.

\subsection*{Acknowledgements}

D. Mas\'is acknowledges the School of Mathematics of the Costa Rica Institute of Technology for their support; this work is part of his M.Sc. dissertation at the University of Costa Rica.
E. Segura and J. Trejos acknowledge the Research Center for Pure and Applied Mathematics (CIMPA) of the University of Costa Rica for their support.
A.E. Xavier acknowledges the Federal University of Rio de Janeiro and the Federal University of Juiz Fora for their support.

\end{document}